\documentclass[10pt,twocolumn,letterpaper]{article}

\usepackage{cvpr}
\usepackage{times}
\usepackage{epsfig}
\usepackage{graphicx}
\usepackage{amsmath}
\usepackage{amssymb}
\usepackage{graphicx}
\usepackage{subfig}
\usepackage{float}
\usepackage{balance}

\usepackage[breaklinks=true,bookmarks=false]{hyperref}

\cvprfinalcopy 

\def\cvprPaperID{****} 

\usepackage[all]{hypcap}
\makeatletter
\newcommand{\printfnsymbol}[1]{%
  \textsuperscript{\@fnsymbol{#1}}%
}
\makeatother
\begin{document}

\title{cGANs for Cartoon to Real-life Images}


\author{Pranjal Singh\\ Rajput\thanks{Denotes equal contribution}\\
\and
Kanya Satis\printfnsymbol{1}\\
\and
Sonnya\\ Dellarosa\printfnsymbol{1}\\
\and
Wenxuan\\ Huang\printfnsymbol{1}\\
\and
Obinna Agba\printfnsymbol{1}\\
}

\affiliation{%
\large{Delft University of Technology} \\
}

\maketitle

\section{Introduction}
Image-to-image translation is a learning task to establish visual mapping between an input and output image. The task has several variations differentiated based on the purpose of the translation, such as synthetic$\xrightarrow{}$real translation~\cite{pan2017virtual}\cite{tsai2019domain}, photo$\xrightarrow{}$caricature translation~\cite{zheng2019unpaired} and many others. The problem has been tackled using different approaches, either through traditional computer vision methods~\cite{hertzmann2001image}, as well as deep learning approaches in recent trends.

One approach currently deemed popular and effective is using conditional generative adversarial network, also known shortly as cGAN~\cite{cgans}. It is adapted to perform image-to-image translation tasks with typically two networks: a generator and a discriminator\cite{Pix2Pix}. The generator attempts to generate a \textit{duplicated imitation} of the input data distribution from a noise distribution while the discriminator classifies whether the generator's input is fake, i.e. imitation from the generated distribution, or real, i.e. ground truth from the original distribution.

Previous research has focused on specific purpose for cross-image translation. Efros et al.~\cite{Efros01} proposed a simple model to synthesize an image based on stitches of input images, which represents a traditional approach directly from input image. Fergus et al.~\cite{Fergus06} addressed specific problems of blurry image into higher crispness, which provided insights on retaining the crispness of duplicated imitation. Additionally, through user studies and an automated mechanism to select images, Chen et al.~\cite{Chen09} developed a model to optimize image selection process before cross-image translation. 

This project is based on an existing implementation of a network called Pix2Pix\cite{Pix2Pix}, a cGAN implementation based on U-Net architecture and convolution Markovian discriminator. The use of U-Net architecture~\cite{Ronneberger15}, instead of the general encoder-decoder architecture is to improve the efficiency of processing input and output with higher resolution while facilitating the transfer of shared, low-level information directly across layers. The U-Net architecture enables cross-layer communication and therefore could more efficiently retain prominent input features, e.g. edge prominence. 

In addition, Pix2Pix also employs a traditional approach, such as L1 distance, to imitate the ground truth and reduce blurring. In this case, L1 loss captures low-frequency crispness level~\cite{LarsenSW15}. The Markovian discriminator is employed to capture high frequency crispness level. It evaluates the image integrity with the unit of patch, namely the PatchGAN, with fewer parameters of image quality required while operating faster~\cite{LiChuan16}. cGANs also combine the input data with the noise distribution to condition the resulting output to the original distribution. Pix2Pix applies this technique for image-to-image translation.

The advantage of this cooperated approach is that the Pix2Pix model can be applied to a varying types and depth of image-to-image translation problems without specifying loss functions specific to the application domain. Pix2Pix has already been applied to different image translation problems with notable success.

\section{Problem statement}
This project aims to evaluate the robustness of the Pix2Pix model by applying the Pix2Pix model to datasets consisting of cartoonized images. Using the Pix2Pix model, it should be possible to train the network to generate real-life images from the cartoonized images. The questions this project seeks to answer are:

\begin{enumerate}
    \item Does the default set of hyper parameters of the original implementation work well on this cartoonized dataset?
    \item Does the model properly recreate facial expressions and postures?
\end{enumerate}

The project's experiment attempts to explore the robustness of the multi-purpose Pix2Pix model. The problem is to determine whether, under the standard hyper-parameters, Pix2Pix can accurately recreate the facial features of the images in the original distribution based on cartoonized version of the images.

\subsection{Dataset}
The dataset selected to train the Pix2Pix model is based on an adaption of the colorFERET~\cite{colorFeret} facial image data. The colorFERET  database contains 11337 images of faces in different angles and postures. These images are taken from 1199 individuals and are typically used for facial recognition research.

The image data is provided in two versions: a higher resolution (512x768) and a lower resolution (256x384) version. In this project, the lower resolution version was used, in attempt to constrain the training duration.

\subsubsection{Data Pre-Processing}
This project utilizes a derivation of the colorFERET images. To obtain the actual data for training and testing the network, a modified version of an OpenCV cartoonizer script~\cite{cartoonizer} was run on selected images to generate their cartoon versions. The pair of  the cartoon and real images are then used as an input to the Pix2Pix model.

\subsubsection{Data Filtering}
After the first evaluation of the network, it was seen that the network seemed to performed badly due to the presence of some intermittent grayscale images, as well as images with low face-to-image ratio, where the person's face only cover a small portion of the full image. These images were later removed through manual inspection to create the filtered dataset. The filtered dataset contains a total of 8933 images which was divided into three parts: 60\% for training, 20\% for testing and 20\% for validation.

\subsubsection{Data Augmentation}
For further evaluation through data augmentation, a set of images were used to create 4 different types of cartoons by varying the parameters of the OpenCV script such as image blurring, gray-scale percentage, etc. A total of 7,858 images were used for this purpose. In order to bring in domain adaptation and test its effect, the equivalent CycleGAN network was used for this purpose.

\subsection{Expected Results} 
Motivated by the many successful applications~\cite{Pix2PixSite} of Pix2Pix with different datasets, it is expected that the Pix2Pix model would perform similarly well, at least on a test data with similar distribution to the training data.

With data augmentation and domain adaptation, the network is expected to perform better for other test data types as well.

\subsection{Evaluation}
One of the challenges cited by the original Pix2Pix paper was a method of evaluating the model. In their examples, they used Amazon Mechanical Turks, a crowd-sourcing marketplace, to evaluate if a person is able to distinguish whether an image is real or fake. For a more quantitative evaluation, they utilized the so-called FCN score, in which the authors trained a fully convolutional network to correctly label the original data with high confidence. This trained network was then used to classify the generated samples with the expectation that it would also accurately classify the generated data with a high confidence.

However, for this project, the result is initially analyzed visually and later we use the cosine dissimilarity to quantitatively evaluate our network.

\section{Technical approach}
This section details the use of cGANs and Pix2Pix, and the training of the network.

\subsection{cGANs}
Generative adversarial networks (GANs) \cite{gans} have been used to train generative models, which can generate data from some input. GANs use an adversarial model where two networks, namely the discriminator $D$ and the generator $G$, play a sort of \textit{minmax} game. The generator network attempts to generate synthetic data from a noise distribution and the discriminator estimates the probability that a data sample was generated from the original data distribution as opposed to the synthetic distribution. Equation~\ref{equation: gan_eqn} below captures the function which this adversarial model tries to optimize.

\begin{equation}
\begin{split}
    \min_{G}\max_{D}V(G,D) = \mathbb{E}_{x \sim p_{data}(x) \lbrack \log{D(x)} \rbrack } \\ 
    + \mathbb{E}_{z \sim p_{z}(z) \lbrack \log{1 - D(G(z))} \rbrack }
\end{split}{}
\label{equation: gan_eqn}
\end{equation}

In Equation~\ref{equation: gan_eqn}, the model aims to maximize the probability that the discriminator $D$ assigns the correct label, whether the image is real or fake, to the input data while also minimizing the probability that data generated by the generator $G$ is classified as real, as denoted by $\log{1 - D(G(z))}$.

cGANs differ from the GAN model described above in that the adversarial model is conditioned using some additional data. This conditioning has the effect on exerting more control, i.e. providing a direction, on the data being generated. Equation~\label{equation:cgan_eqn} details the optimization function for cGANs.

\begin{equation}
\begin{split}
    \min_{G}\max_{D}V(G,D) = \mathbb{E}_{x \sim p_{data}(x) \lbrack \log{D(x \| y)} \rbrack } \\ 
    + \mathbb{E}_{z \sim p_{z}(z) \lbrack \log{1 - D(G(z \| y))} \rbrack }
\end{split}{}
\label{equation:cgan_eqn}
\end{equation}

\subsection{Pix2Pix}
Pix2Pix~\cite{Pix2Pix} uses a cGAN model for image-to-image translation tasks of translating one possible representation of a scene to another. In the case of Pix2Pix, the generator $G$ and discriminator $D$ models are conditioned using the input image itself.

The original Pix2Pix implementation modifies the optimization function of cGANs from Equation~\ref{equation:cgan_eqn} by adding an L1 loss term when training the generator $G$ model. Equation~\ref{equation: Pix2PixGen} describes the optimization function for the generator $G$ in Pix2Pix.

\begin{equation}
    G^{*} = \arg \min_{G} \max_{D} L_{cGAN}(G,D) + \lambda L_{L1}(G)
    \label{equation: Pix2PixGen}
\end{equation}
\\
where $L_{cGAN}(G,D) = \mathbb{E}_{z \sim p_{z}(z) \lbrack \log{1 - D(G(z \| y))} \rbrack }$ \\

and $\lambda L_{L1}(G) = \mathbb{E}_{x,y,z} \lbrack \| y - G(x,z) \| \rbrack$ \\

The L1 loss in this case models the L1 per-pixel difference between the generated images and the ground truth. The authors picked L1 over L2 as the L2 distance produced more blurry results.

The authors also provided an implementation of the Pix2Pix model on GitHub~\cite{gitPix2Pix}. This implementation was originally written in \texttt{lua}. However, this project uses a \texttt{PyTorch}~\cite{pytorchPix2Pix} implementation of the paper.

In this implementation, a U-Net 256 network~\cite{unet} was used as the generator network. For the discriminator, a PatchGAN network proposed by the Pix2Pix authors was used. This network restricts the evaluation scope to patches instead of the full image. In other words, it attempts to classify if each $N\times N$ patch in the image is real or fake. The authors also showed that $N$ can be made much smaller than the image size without affecting the results.

\subsection{Training the Model}
When training the model, the input images were first scaled, cropped and then re-sized to 256x256 pixels. This is to condition the input for the U-Net 256 network which requires a resolution of 256x256 for the input images. Table~\ref{table: train_params} shows the default hyper-parameters that are used and Table~\ref{table: tests} shows all the types of networks that were trained by variation of different hyper-parameters and datasets.

\begin{table}[h]
    \centering
    \begin{tabular}{|c|c|}
    \hline
     Name &  Value \\
     \hline
     Batch Size & 1 \\
     Beta1 & 0.5 \\
     Beta2 & 0.9999 \\
     Discriminator Network & PatchGAN \\
     Epsilon & 1e-08 \\
     Generator Network & U-Net 256 \\
     Learning Rate & 0.0002 \\
     L1 $\lambda$ & 100.0 \\
     Optimization Method & Adam \\
     Weight Decay & 0 \\
     \hline
    \end{tabular}
    \caption{Training Configuration and Hyper-Parameters}
    \label{table: train_params}
\end{table}

\begin{table}[h]
    \begin{tabular}{|c|l|l|l|}
    \hline
    \textbf{No} & \textbf{Network}  & \textbf{Dataset}  & \textbf{Hyper-parameters} \\ \hline
    1       & Pix2Pix  & Original  & No change               \\ \hline
    2       & Pix2Pix  & Filtered  & No change               \\ \hline
    3       & Pix2Pix  & Filtered  & Batch size = 8          \\ \hline
    4       & Pix2Pix  & Filtered  & Batch size = 64         \\ \hline
    5       & Pix2Pix  & Augmented & No change               \\ \hline
    6       & CycleGAN & Filtered  & No change               \\ \hline
    \end{tabular}
    \caption{Variations used for training different models}
    \label{table: tests}
\end{table}

The training was carried out on the Google Cloud platform using an NVIDIA Tesla V100 GPU. It was performed for 200 epochs. The duration of training a Pix2Pix network for the default set of parameters (models 1, 2, 5 in Table~\ref{table: tests}) was approximately 8 hours, for a batch size of 8 (model 3) was 5 hours, for a batch size of 64 (model 4) was 2 hours, and training on a CycleGAN network for the default set of parameters (model 6) took about 10 hours.

\section{Experiments and Results}
This section describes the results of our experiment as we attempted to continuously improve the results by filtering the dataset, changing the hyper-parameters, and using a different network.

\subsection{Initial Run}
\begin{figure}[H]%
    \centering
    \subfloat[Sample result: input, generated, ground truth]{
    {
    \includegraphics[scale=0.25]{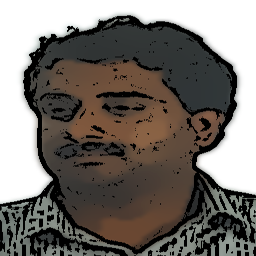} 
    \includegraphics[scale=0.25]{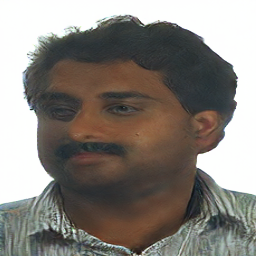}
    \includegraphics[scale=0.25]{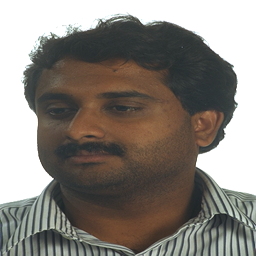} 
    }
    \label{fig:MFLb}
    }%

    \subfloat[Sample result: input, generated, ground truth]{{
    \includegraphics[scale=0.25]{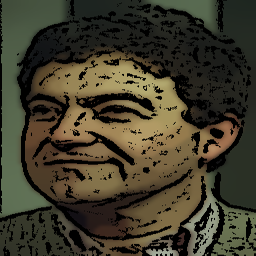}
    \includegraphics[scale=0.25]{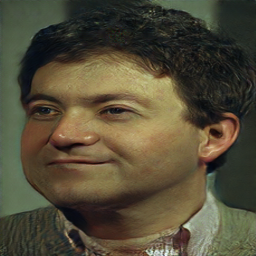}
    \includegraphics[scale=0.25]{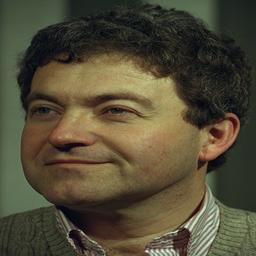}
    }
    \label{fig:MFLc}
    }%
    
    \subfloat[Grayscale image: input, generated, ground truth]{
    {
    \includegraphics[scale=.25]{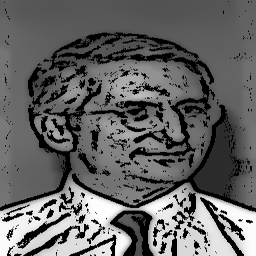} 
    \includegraphics[scale=.25]{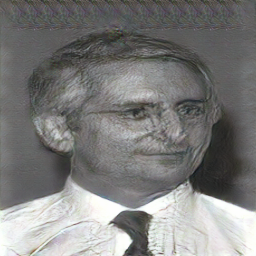}
    \includegraphics[scale=.25]{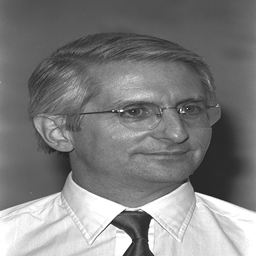}
    }
    \label{fig:MFLa}
    }%
    
    \subfloat[Low face-to-image ratio: input, generated, ground truth]{{
    \includegraphics[scale=0.25]{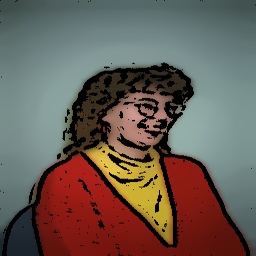} 
    \includegraphics[scale=0.25]{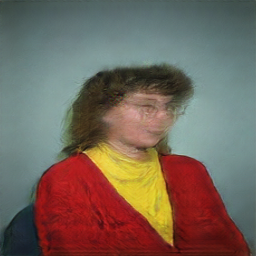}
    \includegraphics[scale=0.25]{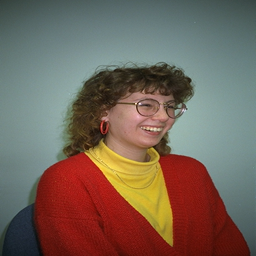}
    }
    \label{fig:MFLd}
    }%
    \caption{Sample results from test on unfiltered dataset}%
    \label{fig:MFL}%
\end{figure}

Using the default hyper-parameter settings shown in Table~\ref{table: train_params}, the Pix2Pix model was trained using the cartoonized data. Upon visual inspection, many generated images resembled the original images. A sample of these can be seen in Figures \ref{fig:MFLb} and \ref{fig:MFLc}.

However, from visual inspection, the performance on the test data was not as good as the one observed during training. It was also seen that the network could not capture the facial expressions in grayscale images properly, resulting in a generated image that appeared blurry, as seen in Figure~\ref{fig:MFLa}. In addition, images with a low face-to-image ratio, in which the face did not occupy a large portion of the image also produced blurry results, as seen in Figure~\ref{fig:MFLd}.

These bad result cases of grayscale images and images with a low face-to-image ratio might be explained by the fact that the images not having enough data for an accurate translation. In the case of the grayscale images, only one channel is available and in the case of images with a low face-to-image size ratio, at the used resolution of 256x256, there simply is not enough data for a proper reconstruction.

\subsection{Training with the Filtered Dataset}
\begin{figure}[ht]%
    \centering
    \subfloat[Sample Cartoon: Input, Generated, Ground Truth]{
    {
    \includegraphics[scale=.25]{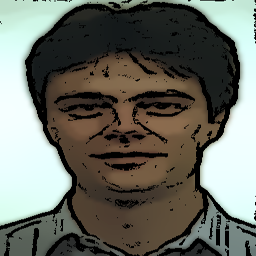} 
    \includegraphics[scale=.25]{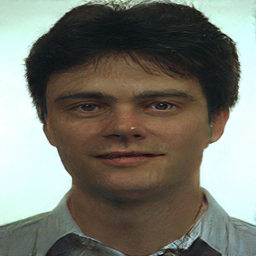}
    \includegraphics[scale=.25]{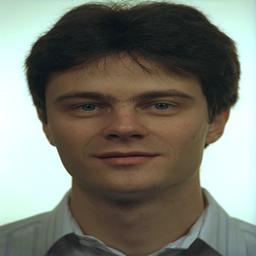}
    }
    \label{fig:batch1Filtereda}
    }%
    
    \subfloat[Sample Cartoon: Input, Generated, Ground Truth]{
    {
    \includegraphics[scale=.25]{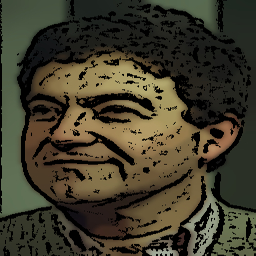} 
    \includegraphics[scale=.25]{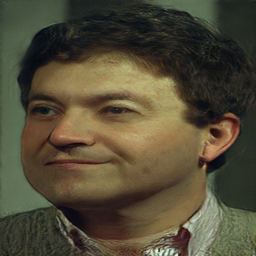}
    \includegraphics[scale=.25]{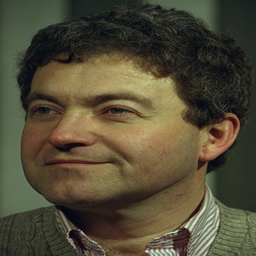}
    }
    \label{fig:batch1Filteredb}
    }%
    
    \subfloat[Sample Cartoon: Input, Generated, Ground Truth]{
    {
    \includegraphics[scale=.25]{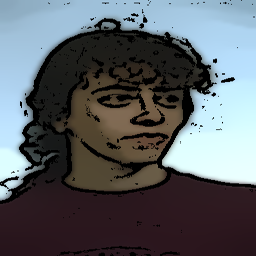} 
    \includegraphics[scale=.25]{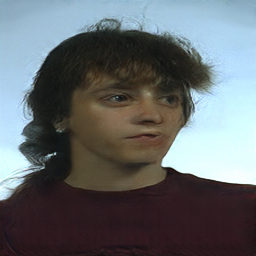}
    \includegraphics[scale=.25]{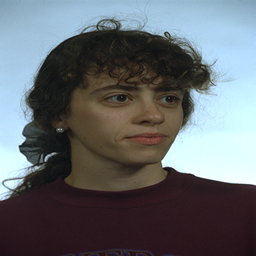}
    }
    \label{fig:batch1Filteredc}
    }%
    \caption{Sample results from test on filtered dataset}%
    \label{fig:batch1Filtered}%
\end{figure}

In a bid to obtain a better trained model, grayscale images and images with a low face-to-image ratio were filtered out. The Pix2Pix model was then retrained on this filtered dataset. Sample results from this run are shown in Figure~\ref{fig:batch1Filtered}. Visually, there was no apparent improvement over the run with the unfiltered dataset. This only removed the blurry results.

\subsection{Varying the Batch Size}
\begin{figure}[ht]%
    \centering
    \subfloat[Sample Cartoon: Input, Generated, Ground Truth]{
    {
    \includegraphics[scale=.25]{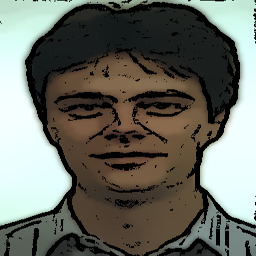} 
    \includegraphics[scale=.25]{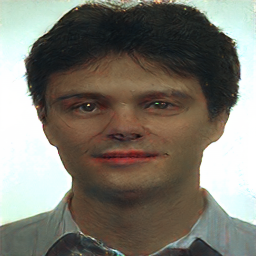}
    \includegraphics[scale=.25]{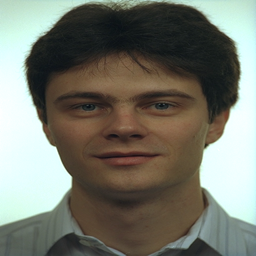}
    }
    \label{fig:batch8Filtereda}
    }%
    
    \subfloat[Sample Cartoon: Input, Generated, Ground Truth]{
    {
    \includegraphics[scale=.25]{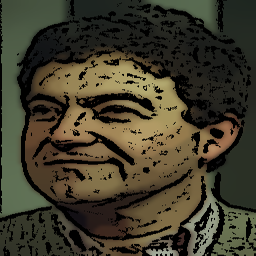} 
    \includegraphics[scale=.25]{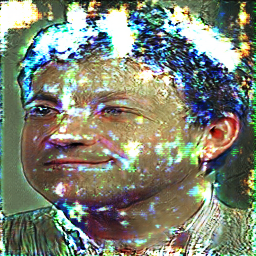}
    \includegraphics[scale=.25]{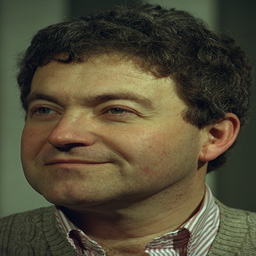}
    }
    \label{fig:batch8Filteredb}
    }%
    
    \subfloat[Sample Cartoon: Input, Generated, Ground Truth]{
    {
    \includegraphics[scale=.25]{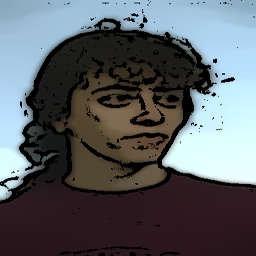} 
    \includegraphics[scale=.25]{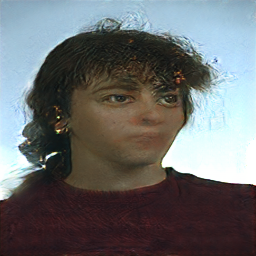}
    \includegraphics[scale=.25]{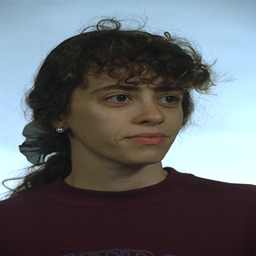}
    }
    \label{fig:batch8Filteredc}
    }%
    \caption{Sample results from test on filtered dataset with batch size 8}%
    \label{fig:batch8Filtered}%
\end{figure}

\begin{figure}[ht]%
    \centering
    \subfloat[Sample Cartoon: Input, Generated, Ground Truth]{
    {
    \includegraphics[scale=.25]{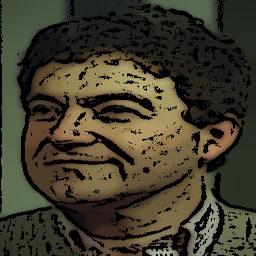} 
    \includegraphics[scale=.25]{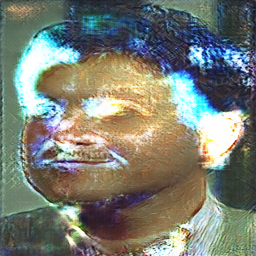}
    \includegraphics[scale=.25]{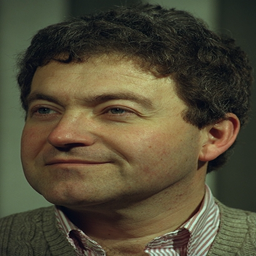}
    }
    \label{fig:batch64Filtereda}
    }%
    
    \subfloat[Sample Cartoon: Input, Generated, Ground Truth]{
    {
    \includegraphics[scale=.25]{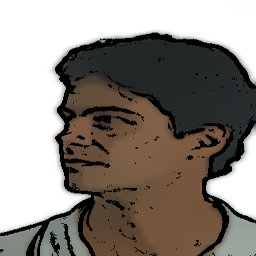} 
    \includegraphics[scale=.25]{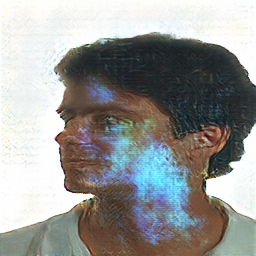}
    \includegraphics[scale=.25]{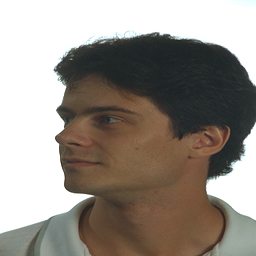}
    }
    \label{fig:batch64Filteredb}
    }%
    
    \subfloat[Sample Cartoon: Input, Generated, Ground Truth]{
    {
    \includegraphics[scale=.25]{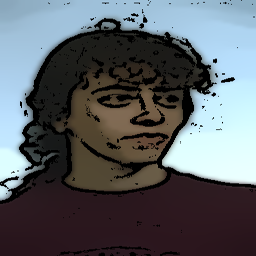} 
    \includegraphics[scale=.25]{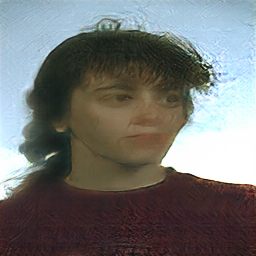}
    \includegraphics[scale=.25]{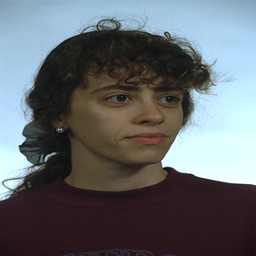}
    }
    \label{fig:batch64Filteredc}
    }%
    \caption{Sample results from test on filtered dataset with batch size 64}%
    \label{fig:batch64Filtered}%
\end{figure}

\begin{figure}[ht]%
    \centering
    {
    {
    \subfloat[Discriminator Loss (Fakes)]{\includegraphics[scale=.25] {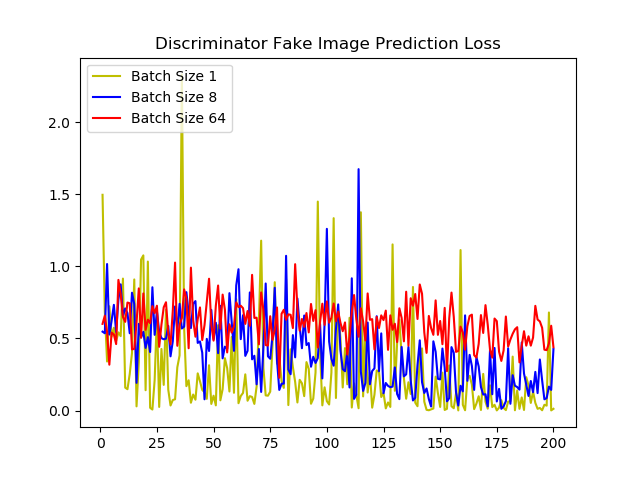}
    \label{fig:lossPlotsa}}
    \subfloat[Discriminator Loss (Real)]{\includegraphics[scale=.25] {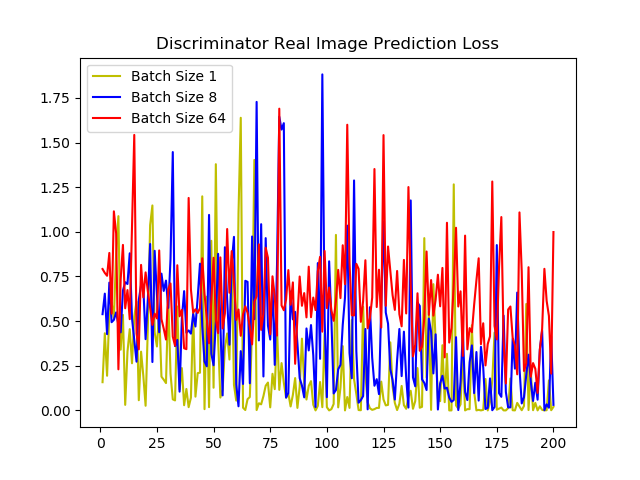}
    \label{fig:lossPlotsb}}
    
    \subfloat[Generator GAN Loss]{\includegraphics[scale=.25] {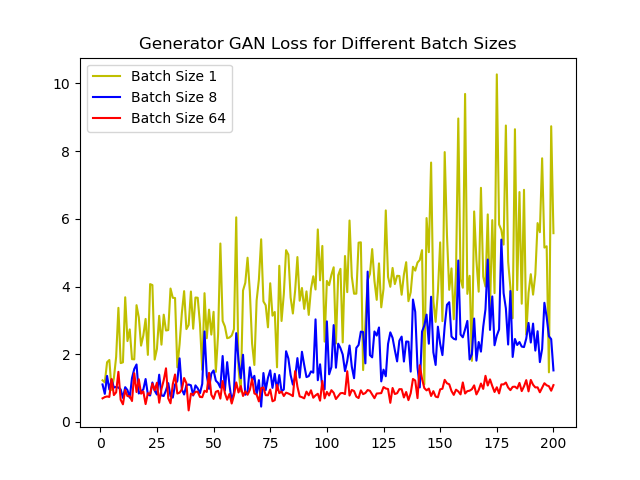}
    \label{fig:lossPlotsc}}
    \subfloat[Generator L1 Loss]{\includegraphics[scale=.25] {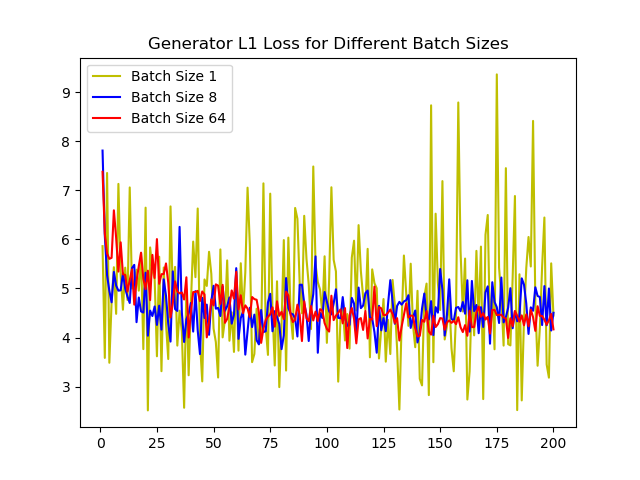}
    \label{fig:lossPlotsd}
    }}
    }%
    \caption{Loss graphs of the different batch sizes}%
    \label{fig:lossPlots}%
\end{figure}

The previous experiments answered the question of how well the default set of hyper-parameters transfers to an application domain, different to those featured in the original paper. Due to the size of the dataset, the training time was quite large, 8 hours on the NVIDIA Tesla V100. To speed up the training time and see its effects on the quality of the results, the batch size was varied. The model was trained using a batch size of 8 and 64. The results from this run are shown in Figure~\ref{fig:batch8Filtered} and Figure~\ref{fig:batch64Filtered} respectively. It is seen that as the batch size increases, the visual quality of the output decreases. Figure~\ref{fig:lossPlots} compares the training loss curves for the generator and discriminator with different batch sizes.

In Figures \ref{fig:lossPlotsc} and \ref{fig:lossPlotsd}, it can be seen that the generator GAN loss is both more stable and lower for higher batch sizes. The generator L1 loss is also more stable as the batch size increases. However, the inverse is the case for the discriminator. The discriminator loss for distinguishing between real and fake images is more unstable and on average higher as the batch size is increased. This could be interpreted to mean that at higher batch sizes, the discriminator is not as effective as it is with a batch size of 1. This would also translate to the generator appearing to perform better at higher batch sizes, since it is not being penalized well enough by its discriminator. This would explain why the visual quality of the generated images at the batch size of 1 is better than that generated by a higher batch size.

It was also observed that for a higher batch size, the L1 loss is higher than the GAN loss, a possible indicator that increasing the batch size should be accompanied with a change in other hyper-parameters.

\begin{figure}[htb]%
    \centering
    {
    {
    \subfloat[Input]{\includegraphics[scale=.3] {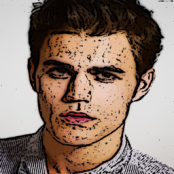}}
    \subfloat[Batch 1]{\includegraphics[scale=.3] {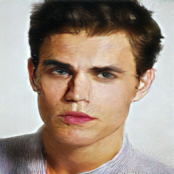}}
    \subfloat[Batch 8]{\includegraphics[scale=.407] {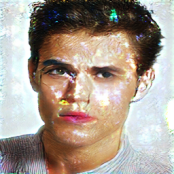}}
    \subfloat[Batch 64]{\includegraphics[scale=.407] {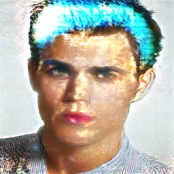}}
    \subfloat[Gr. truth]{\includegraphics[scale=.3] {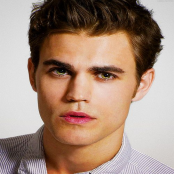}
    }}
    }%
    \caption{Sample results on various batch sizes}%
    \label{fig:oodSample1}%
\end{figure}

\subsection{Data Augmentation}
\begin{figure}[ht]
    \centering
    \subfloat[Style 1: Cartoon, Generated, Ground Truth]{
    {\includegraphics[scale=.30]{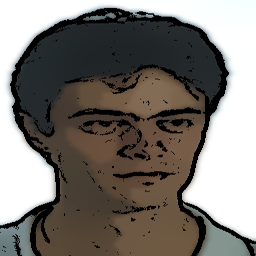} }
    {\includegraphics[scale=.30]{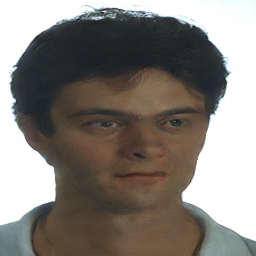} }
    {\includegraphics[scale=.30]{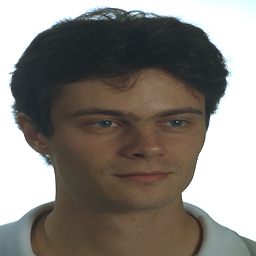} }
    }%
    
    \subfloat[Style 2: Cartoon, Generated, Ground Truth]{
    {\includegraphics[scale=.30]{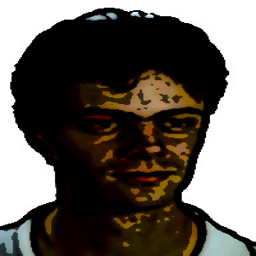} }
    {\includegraphics[scale=.30]{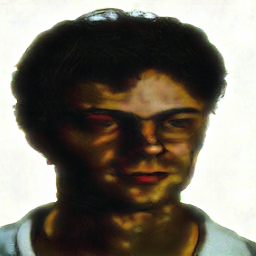} }
    {\includegraphics[scale=.30]{images/cartoonizer_real.png} }
    }%
    \caption{Training with different cartoonization techniques}%
    \label{fig:adversarialTraining}%
    
\end{figure}

The next question which needed to be answered was how well the model performs on out-of-distribution (OOD) samples. One technique to improve performance on such distributions involves constructing a common representative space~\cite{domainAdaptation1}~\cite{domainAdaptation2}. To achieve this, the script which was used to generate the synthetic data was slightly modified to generate different style of cartoons for the same image. This allowed different samples to map to the same distribution. Figure~\ref{fig:adversarialTraining} shows for the same image with different cartoonization techniques and the corresponding generated image
Figure~\ref{fig:oodSample1} shows the performance of the model for different batch sizes, after training on this augmented dataset, on such an OOD sample.

\begin{figure}[ht]%
    \centering
    \subfloat[Sketch Input]{{\includegraphics[width=.15\textwidth]{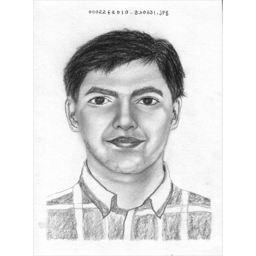} }}%
    \hfill
    \subfloat[Generated Image]{{\includegraphics[width=.15\textwidth]{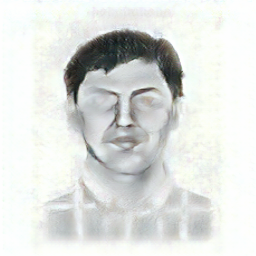} }
    \label{fig:sketch_real_b}}%
    \hfill
    \subfloat[Ground Truth]{{\includegraphics[width=.15\textwidth]{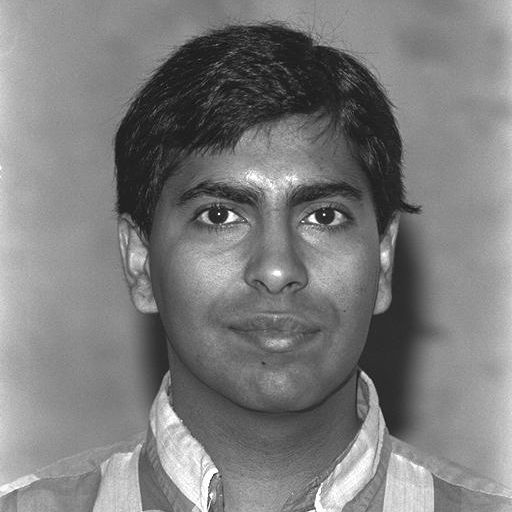} }}%
    \caption{Image samples in case of hand drawn sketches (an OOD sample)}%
    \label{fig:sketch_real}%

\end{figure}

The downside with this technique, however, is that the higher the difference between the OOD sample and the adversarial samples on which the network is trained the lower the quality of the generated image. Figure~\ref{fig:sketch_real} shows the generated image when the input is a sketch of the original image.

\subsection{Comparison to CycleGAN}
The authors of Pix2Pix later attempted in improving the quality of image-to-image translation by building a state-of-the-art network called CycleGAN~\cite{DBLP:journals/corr/ZhuPIE17} that builds on Pix2Pix. The main difference with Pix2Pix is that CycleGAN learns the mapping without paired training examples. CycleGAN learns a mapping $G: X \xrightarrow{} Y$ such that the distribution of images from $G(X)$ is highly identical to the distribution $Y$ using an adversarial loss. As this mapping is highly under-constrained, it is coupled with an inverse mapping $F: Y \xrightarrow{} X$ to introduce a cycle consistency loss such that $F(G(X)) \approx X$ and vice versa.

\begin{figure}[ht]%
    \centering
    {
    {
    \subfloat[Input]{\includegraphics[scale=.7] {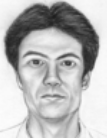}
    \label{fig:Pix2Pixvscyclegan_a}
    }
    \subfloat[Pix2Pix]{\includegraphics[scale=.7] {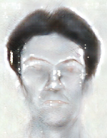}
    \label{fig:Pix2Pixvscyclegan_b}}
    \subfloat[CycleGAN]{\includegraphics[scale=.7] {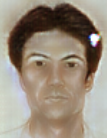}
    \label{fig:Pix2Pixvscyclegan_c}}
    \subfloat[Ground truth]{\includegraphics[scale=.7] {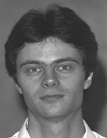}
    \label{fig:Pix2Pixvscyclegan_d}}
    }
    }%
    \caption{Sample results on Pix2Pix vs CycleGAN}%
    \label{fig:Pix2Pixvscyclegan}%
\end{figure}

We trained our dataset on the CycleGAN framework and compared its test results, shown in Figure~\ref{fig:Pix2Pixvscyclegan_b}, with those of Pix2Pix, Figure~\ref{fig:Pix2Pixvscyclegan_c}. It can be observed here that CycleGAN outputs a much better image. CycleGAN does not remove the person's eyes and is able to approximate the color of the person's skin. 

\subsection{Quantitative Evaluation of Generated Images}
Up to this point, the quality of the results has been reported only through a visual inspection. Quantitatively evaluating the generated results for GAN models is still an ongoing research area. The original paper utilized the so-called FCN Score. However, the network used to estimate the FCN score is especially suited for segmentation classification. In this report, two other evaluation techniques are utilized.

\subsubsection{Cosine Dissimilarity}
Cosine dissimilarity is a measure of the similarity between two vectors. Since the generated image and the target image are themselves representable as vectors, it is therefore possible to use the cosine dissimilarity to measure how close the two images are. Equation~\ref{equation:cosine_dissimilarity} gives a general formula for computing the cosine similarity between two vectors.

\begin{equation}
    similarity = \cos{\theta} = \frac{\mathbf{A} \cdot \mathbf{B}}{\| \mathbf{A} \|\| \mathbf{B} \|}
    \label{equation:cosine_dissimilarity}
\end{equation}{}

This measure does not penalize really small pixel differences, as would be the case in per-pixel comparison of the L1 distance. To account for negative values of $\cos{\theta}$, Equation~\ref{equation:cosine_dissimilarity} is modified slightly to become Equation~\ref{equation:cosine_dissimilarity2}

\begin{equation}
    similarity = 1 - \cos{\theta} = 1 - \frac{\mathbf{A} \cdot \mathbf{B}}{\| \mathbf{A} \|\| \mathbf{B} \|}
    \label{equation:cosine_dissimilarity2}
\end{equation}{}

\begin{figure}[ht]
    \includegraphics[scale=0.5] {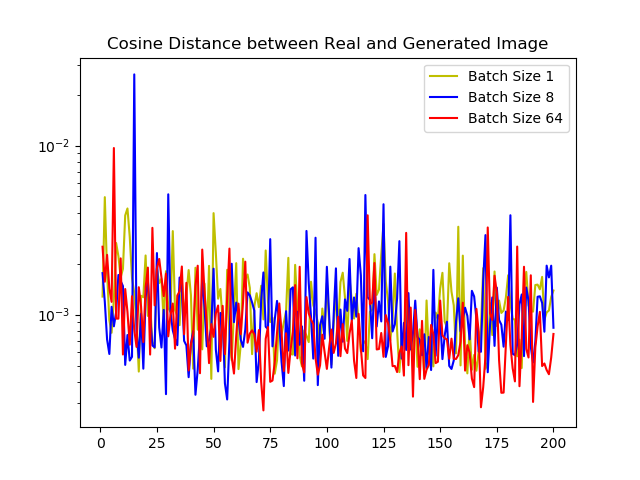}
    \caption{Cosine Similarity during training with different Batch Sizes}%
    \label{fig:cosine_distance_batchsize}%
\end{figure}

In this new form, the values range from [0, 2] with a higher score indicating a higher level of dissimilarity. Figure \ref{fig:cosine_distance_batchsize} compares the similarity measure during training for the different batch sizes.

\begin{table}[h]
    \centering
    \begin{tabular}{|c|c|}
    \hline
    \textbf{Batch Size}  & \textbf{Mean Cosine Dissimilarity}  \\ \hline
    1       & 0.00136 \\ \hline
    8       & 0.00622 \\ \hline
    64       & 0.01052 \\ \hline
    \end{tabular}
    \caption{Mean Cosine Dissimilarity for different Batch Sizes}
    \label{table:mean_cosine_similarity}
\end{table}

From Figure \ref{fig:cosine_distance_batchsize}, it can be seen that on average, the cosine dissimilarity between the different batch sizes is the same. However, in line with the observations made through a visual inspection, Table~\ref{table:mean_cosine_similarity} shows that the model trained with a batch size of 1 has the lowest cosine value and that the higher the batch size the worse the score. The disparity between this observation and that obtained when comparing the dissimilarity during training suggests that the higher batch size models do not generalize as well as the model with a batch size of 1.

\subsubsection{Probability Score of an Age and Gender CNN}
As an alternative to the FCN score, results were evaluated using a convolution neural network (CNN) trained to classify the gender of a facial image. A pre-trained Inception V3 model~\cite{inceptionV3} was used as the classifier for this evaluation. For high quality generated images, the classifier should assign the correct label to the generated image with a similar level of confidence as it did to the original image. The gender label was used because the original FERET database was nearly evenly split between male and female. Other provided labels such as age and race did not provide as even a distribution.

The pre-trained inception model had an accuracy of 96\% on the dataset used for training. Table \ref{table: inception_score} shows the mean absolute difference between predictions for the original images and the generated images for the test data for different batch sizes.

Results in Table \ref{table: inception_score} do not yield the same observation as the cosine similarity and the visual inspection evaluations, with the batch size 8 model having the least difference between the prediction probabilities for real and generated images. A possible explanation for this might be that gender classification is perhaps not the best label for evaluating the classification accuracy.

\begin{table}[h]
    \centering
    \begin{tabular}{|c|c|}
    \hline
    \textbf{Batch Size}  & \textbf{Mean Absolute Difference}  \\ \hline
    1       & 0.075 \\ \hline
    8       & 0.060 \\ \hline
    64       & 0.078 \\ \hline
    \end{tabular}
    \caption{Mean absolute difference of image prediction probabilities for different batch sizes}
    \label{table: inception_score}
\end{table}

\section{Discussions}
In this section, we discuss important aspects that greatly affected the results of our experiment.

\subsection{Effect of batch sizes}
Our finding that a higher batch size, while significantly reduces computation time, negatively affects the resulting generated image is reflected in existing literature. In practice, it has been seen that using a larger batch size results in a significant degradation in the quality of the model, as measured by its ability to generalize~\cite{DBLP:journals/corr/KeskarMNST16}. This lack of ability to generalize is mostly due to the fact that methods with high batch sizes tend to converge to sharp minimizers of the training function. Mishkin et al.~\cite{DBLP:journals/corr/MishkinSM16} have also shown, in their studies on the performance of a CNN, that large mini-batch sizes results in a worse accuracy.

\subsection{Downsides of Data Generation Techniques}
The cartoons used in the training were unfortunately not real cartoons, but rather they were generated using OpenCV, which essentially added a filter to the image where the edges in the image are traced and the colors are simplified to resemble a cartoon picture. As a result, what the network learned was actually the features of this image filter and not the underlying features of image-to-image translation. This is evident when testing the network using an actual hand-drawn sketch, as seen in Figure~\ref{fig:sketch_real_b}, where the trained Pix2Pix network outputs an image with the edges removed and some areas that are slightly blurred.

\section{Conclusion}
We have performed an experiment with several test rounds in order to evaluate the general robustness and applicability of Pix2Pix as solution to image-to-image translation tasks. On the test run where default parameters and configuration are retained, the result is less satisfying for images with non-amplified features, such as the lack of color or the lack of prominent facial features. This suggests that the model is susceptible to instability when the diversity of the image datasets varies. 

In a later stage of the project, several alterations on data synthesis and hyper-parameters Were introduced. Image filtering assured a higher homogeneity within the distribution of the dataset, while the positive effect, in terms of computation time, of increasing the batch size is superseded by its damping effect on the discriminator. The performance of Pix2Pix on out-of-distribution samples also differs on the degree of variation between the original sample and the adversarial samples, which indicates unstable quality assurance under visual changes. 

Moreover, a better evaluation methodology is required. The current methods employed can be easily categorized as subjective as the result and degree of variation from qualitative comparisons differs per individuals.

In general, the result of this report indicates acceptable robustness over varying image types and qualities. In case where images are restricted with standard color and prominent facial features, the Pix2Pix model is a reliable image-to-image translation solution.

{\small
\bibliographystyle{ieee_fullname}
\bibliography{egbib}
}

\appendix







    


\end{document}